\pdfoutput=1
\documentclass[conference]{IEEEtran}

\usepackage{cite}
\usepackage{amsmath}
\usepackage{graphicx}
\usepackage{booktabs}
\usepackage{array}
\usepackage{siunitx}
\usepackage{xcolor}
\usepackage{tikz}
\usetikzlibrary{positioning,arrows.meta,fit,calc}
\usepackage[hidelinks]{hyperref}
\hypersetup{pdftitle={Point-in-Time Encoders for Earnings Calls}, pdfauthor={Alessio Brini}}
\newcommand{\AblLate}{\num[round-mode=places,round-precision=4]{0.006547619047619048}}
\newcommand{\AvBaseIC}{\num[round-mode=places,round-precision=3]{0.5129295864047784}}
\newcommand{\BonferroniBound}{\num[round-mode=places,round-precision=2]{3.4698065551615747}}
\newcommand{\BonferroniBoundStudent}{\num[round-mode=places,round-precision=2]{3.72626675050224}}
\newcommand{\BreakEvenBase}{\num[round-mode=places,round-precision=0]{13.563686037766818}}
\newcommand{\BreakEvenOtcOurs}{\num[round-mode=places,round-precision=0]{23.39131763090604}}
\newcommand{\BreakEvenOurs}{\num[round-mode=places,round-precision=0]{11.871886479626934}}
\newcommand{\ChainLate}{\num[round-mode=places,round-precision=4]{0.008976190476190476}}
\newcommand{\ChainMinusScratch}{\num[round-mode=places,round-precision=3]{0.006928571428571429}}
\newcommand{\ChainSameYear}{\num[round-mode=places,round-precision=4]{0.004119047619047619}}
\newcommand{\ChainSameYearHi}{\num[round-mode=places,round-precision=4]{0.006489817398348465}}
\newcommand{\ChainSameYearLo}{\num[round-mode=places,round-precision=4]{0.0017111081626208665}}
\newcommand{\ChainSameYearPos}{\num[round-mode=none,group-separator={,}]{11}}
\newcommand{\ChanceWithin}{\num[round-mode=places,round-precision=2]{0.05110926621762123}}
\newcommand{\DaptMinusAbl}{\num[round-mode=places,round-precision=3]{0.02766666666666667}}
\newcommand{\KnnSueOurs}{\num[round-mode=places,round-precision=3]{0.047874162876594725}}
\newcommand{\KnnSueOursT}{\num[round-mode=places,round-precision=1]{3.254923433165685}}
\newcommand{\NBonferroni}{\num[round-mode=none,group-separator={,}]{4}}
\newcommand{\NBonferroniBlock}{\num[round-mode=none,group-separator={,}]{2}}
\newcommand{\NBonferroniStudent}{\num[round-mode=none,group-separator={,}]{2}}
\newcommand{\NCallsIndex}{\num[round-mode=none,group-separator={,}]{412248}}
\newcommand{\NForecastPremia}{\num[round-mode=none,group-separator={,}]{96}}
\newcommand{\NextDayBaseIC}{\num[round-mode=places,round-precision=3]{0.02509694423540668}}
\newcommand{\NextDayIncOurs}{\num[round-mode=places,round-precision=3]{0.011838097802667544}}
\newcommand{\NextDayIncOursT}{\num[round-mode=places,round-precision=1]{3.021008615997043}}
\newcommand{\NextDayShareTraded}{\num[round-mode=places,round-precision=0]{40.66646096883677}}
\newcommand{\NextDaySharpeBase}{\num[round-mode=places,round-precision=2]{1.7305737787357636}}
\newcommand{\OtcShareTraded}{\num[round-mode=places,round-precision=0]{23.441860465116278}}
\newcommand{\OursMinusAbl}{\num[round-mode=places,round-precision=3]{0.03126666666666667}}
\newcommand{\OursMinusBestCross}{\num[round-mode=places,round-precision=3]{0.12331111111111111}}
\newcommand{\OursMinusEfive}{\num[round-mode=places,round-precision=3]{0.14413333333333334}}
\newcommand{\OursMinusEfiveCross}{\num[round-mode=places,round-precision=3]{0.1374888888888889}}
\newcommand{\ParamsAbl}{\num[round-mode=places,round-precision=1]{149.014272}}
\newcommand{\ParamsOurs}{\num[round-mode=places,round-precision=1]{33.826176}}
\newcommand{\PoolMedianHi}{\num[round-mode=none,group-separator={,}]{27}}
\newcommand{\PoolMedianLo}{\num[round-mode=none,group-separator={,}]{21}}
\newcommand{\RagChain}{\num[round-mode=places,round-precision=4]{0.07641129126772285}}
\newcommand{\RagChainHi}{\num[round-mode=places,round-precision=4]{0.09415193849209694}}
\newcommand{\RagChainLo}{\num[round-mode=places,round-precision=4]{0.05682437319601895}}
\newcommand{\RagChainPos}{\num[round-mode=none,group-separator={,}]{13}}
\newcommand{\RagClosedLevel}{\num[round-mode=places,round-precision=2]{3.617651783823967}}
\newcommand{\RagGoldChainLate}{\num[round-mode=places,round-precision=1]{65.80714285714286}}
\newcommand{\RagGoldChainPit}{\num[round-mode=places,round-precision=1]{63.54999999999998}}
\newcommand{\RagNoGold}{\num[round-mode=places,round-precision=4]{0.001846001225390605}}
\newcommand{\RagPitLevel}{\num[round-mode=places,round-precision=2]{1.4442286885827778}}
\newcommand{\RecallBgeSmall}{\num[round-mode=places,round-precision=3]{0.5384444444444444}}
\newcommand{\RecallChrono}{\num[round-mode=places,round-precision=3]{0.20033333333333336}}
\newcommand{\RecallEfive}{\num[round-mode=places,round-precision=3]{0.5837111111111111}}
\newcommand{\RecallFinbert}{\num[round-mode=places,round-precision=3]{0.22564444444444445}}
\newcommand{\RecallFiveEfive}{\num[round-mode=places,round-precision=3]{0.8797111111111112}}
\newcommand{\RecallFiveOurs}{\num[round-mode=places,round-precision=3]{0.9517111111111111}}
\newcommand{\RecallOurs}{\num[round-mode=places,round-precision=3]{0.7278444444444444}}
\newcommand{\ReplayCompute}{\num[round-mode=places,round-precision=4]{-0.002738095238095238}}
\newcommand{\ReplayComputeWins}{\num[round-mode=none,group-separator={,}]{3}}
\newcommand{\ReplayLater}{\num[round-mode=places,round-precision=4]{0.006857142857142859}}
\newcommand{\ReplayLaterHi}{\num[round-mode=places,round-precision=4]{0.009204339440524395}}
\newcommand{\ReplayLaterLo}{\num[round-mode=places,round-precision=4]{0.004495725402980975}}
\newcommand{\ReplayLaterWins}{\num[round-mode=none,group-separator={,}]{13}}
\newcommand{\RestateLeakyGap}{\num[round-mode=places,round-precision=4]{0.0055}}
\newcommand{\RestateLeakyGapHi}{\num[round-mode=places,round-precision=4]{0.013348887641067042}}
\newcommand{\RestateLeakyGapLo}{\num[round-mode=places,round-precision=4]{-0.002286740183463871}}
\newcommand{\RestateLeakyLow}{\num[round-mode=places,round-precision=4]{0.005928571428571429}}
\newcommand{\RestateLeakyLowHi}{\num[round-mode=places,round-precision=4]{0.011200069928671787}}
\newcommand{\RestateLeakyLowLo}{\num[round-mode=places,round-precision=4]{0.0008042350787976784}}
\newcommand{\ScratchLate}{\num[round-mode=places,round-precision=4]{-0.00280952380952381}}
\newcommand{\ScratchSameYear}{\num[round-mode=places,round-precision=4]{-0.00028571428571428546}}
\newcommand{\SeedRangeVintage}{\num[round-mode=places,round-precision=3]{0.006333333333333413}}
\newcommand{\StaleChainTwentyFour}{\num[round-mode=places,round-precision=3]{0.7513333333333333}}
\newcommand{\StaleChainTwentyTen}{\num[round-mode=places,round-precision=3]{0.04033333333333333}}
\newcommand{\StaleSlopeChain}{\num[round-mode=places,round-precision=4]{0.002524542124542124}}
\newcommand{\StaleSlopeScratch}{\num[round-mode=places,round-precision=4]{0.002011721611721612}}
\newcommand{\StaleTwiceTwentyFour}{\num[round-mode=places,round-precision=3]{0.756}}
\newcommand{\SueAloneOursHeld}{\num[round-mode=places,round-precision=4]{0.012599096683871316}}
\newcommand{\SueAloneOursHeldHi}{\num[round-mode=places,round-precision=4]{0.027591032844578686}}
\newcommand{\SueAloneOursHeldLo}{\num[round-mode=places,round-precision=4]{-0.010468340628077118}}
\newcommand{\SueAloneOursIC}{\num[round-mode=places,round-precision=3]{0.10109215451660114}}
\newcommand{\SueAloneOursTrained}{\num[round-mode=places,round-precision=4]{0.013533370156730284}}
\newcommand{\SueBaseIC}{\num[round-mode=places,round-precision=3]{0.25383468264020564}}
\newcommand{\SueHeldPerQuarter}{\num[round-mode=places,round-precision=0]{46.645833333333336}}
\newcommand{\SueLateChainAlone}{\num[round-mode=places,round-precision=4]{0.01554344931208002}}
\newcommand{\SueLateChainAloneBlockT}{\num[round-mode=places,round-precision=1]{5.427796540630767}}
\newcommand{\SueLateChainAloneT}{\num[round-mode=places,round-precision=1]{6.067628006758187}}
\newcommand{\SueLateOurs}{\num[round-mode=places,round-precision=4]{0.00444227939304653}}
\newcommand{\SueLateOursAlone}{\num[round-mode=places,round-precision=4]{0.013365469497059551}}
\newcommand{\SueLateOursAloneBlockT}{\num[round-mode=places,round-precision=1]{3.574794594188621}}
\newcommand{\SueLateOursAloneT}{\num[round-mode=places,round-precision=1]{3.7009389657269685}}
\newcommand{\SueLateOursT}{\num[round-mode=places,round-precision=1]{2.695029064820451}}
\newcommand{\SueWithOursIC}{\num[round-mode=places,round-precision=3]{0.24109688589960063}}
\newcommand{\TaskStageGain}{\num[round-mode=places,round-precision=3]{0.5587777777777778}}
\newcommand{\TokensTwentyTen}{\num[round-mode=places,round-precision=1]{0.302647296}}
\newcommand{\TwiceGain}{\num[round-mode=places,round-precision=3]{0.018088888888888887}}
\newcommand{\VersionLateMax}{\num[round-mode=places,round-precision=2]{1.3941381490904279}}
\newcommand{\VersionLateMin}{\num[round-mode=places,round-precision=2]{0.3179396026312756}}
\newcommand{\VersionLateTen}{\num[round-mode=places,round-precision=2]{6.684182275334067}}
 
\newcommand{\CodeURL}{https://github.com/Alessiobrini/point-in-time-encoders-for-earnings-calls}

\begin{document}

\title{Point-in-Time Encoders for Earnings Calls}

\author{\IEEEauthorblockN{Alessio Brini}
\IEEEauthorblockA{\textit{Pratt School of Engineering} \\
\textit{Duke University}\\
Durham, NC, USA \\
alessio.brini@duke.edu}
}

\maketitle

\begin{abstract}
Text encoders turn earnings calls into vectors for search, for language models that answer from retrieved passages, and for forecasting models. A live retriever can use an encoder trained once on all available text, as finance retrievers are, but a backtest needs an encoder that read no text written after the date of each vector. To address this look-ahead bias, from \NCallsIndex{} calls we train vintages under four pretraining recipes, with yearly cutoffs from 2010 to 2024, and each vintage reads only text dated on or before its cutoff. We then fine-tune each vintage on the question-answer pairs of the calls themselves. The vintages we train on calls alone, with \ParamsOurs M parameters, retrieve answers on held-out calls better than the general retrievers we tested in every test quarter. Their vectors also raise by about half the rank correlation between next-day returns and forecasts built from past returns and firm size. When each vintage continues the previous one, a later vintage retrieves better on earlier quarters than the vintage available at the time, and a point-in-time language model that answers from the passages the later vintage retrieves inherits the gain. The 2024 vintage of our families also forecasts earnings surprises better, although the interval of this gain includes zero on calls that no vintage read. It forecasts returns no better, so point-in-time vintages matter most for retrieval and little for forecasts of returns.
\end{abstract}

\begin{IEEEkeywords}
Point-in-time models, look-ahead bias, dense retrieval, contrastive learning, earnings calls, return forecasting
\end{IEEEkeywords}

\section{Introduction}\label{sec:intro}
\looseness=-1 A retriever returns the passages of a corpus most relevant to a query, and its encoder maps each passage and each query to a vector. Over earnings calls, a retriever lets an analyst search hundreds of thousands of calls for statements worded differently from the query. It also supplies the passages from which a language model answers, so the quality of retrieval caps the quality of the answer, and the vectors of its encoder can enter a forecasting model directly. When the retrieved passages only inform a live reader, the retriever can use an encoder trained on the whole corpus, because every text it read was written before the query. Many uses instead treat the passages or the vectors as what was known at a past date, as in a backtest of a forecasting model, a strategy or an agent. Every vector used at date $d$ must then come from an encoder that stopped reading before $d$. For example,~\cite{foley2026word} build topic factors for asset pricing from news with a pretrained sentence encoder that has seen text from a period overlapping their evaluation sample. Its vectors may therefore encode information that was not available on the date of an article. For a language model,~\cite{glasserman2024assessing} note that a backtest produces biased results if the training and backtesting periods overlap, with look-ahead bias as one source. An encoder that read later text is open to the same bias.

\looseness=-1 Point-in-time language models remove look-ahead bias from the language model itself by training a sequence of models, called vintages, each only on text dated before its cutoff. ChronoBERT and ChronoGPT~\cite{he2026chronologically}, DatedGPT~\cite{yan2026datedgpt} and the point-in-time models of~\cite{kelly2026scaling} train on general text and judge their vintages through portfolios built on the models' signals. For decoder signals on next-day returns from news,~\cite{lee2026palm} find no evidence that a newer vintage performs better than the one it replaced on the same evaluation window, and argue that the annual pretraining run is not necessary. However, a dated language model protects a backtest only if the encoder that retrieves or embeds its input is dated as well, and these vintages are not trained to retrieve. Moreover,~\cite{kong2026evaluating} identify retrieval from an external knowledge base as a second source of look-ahead bias, after the model's own training data, and call for point-in-time data and version-controlled corpora with explicit as-of metadata. Similarly,~\cite{zhu2026artificial} include embedding models among the components that must be versioned, since each can encode later information.

To our knowledge, the leak through the encoder of a retriever has not been measured, although finance retrievers exist~\cite{anderson2024greenback,tang2025finmteb,brenner2025adaptation}. Each fine-tunes a general encoder, whose training cutoff is unknown, on queries or relevance labels at least partly produced by a language model, and no paper among them states the cutoff of its encoder.

\looseness=-1 A point-in-time retriever for earnings calls faces two obstacles. Calls before 2010 are few, so a vintage with a 2010 cutoff can read only \TokensTwentyTen B tokens of calls, while a general encoder pretrains on hundreds of billions of tokens. Scoring each yearly vintage also requires retrieval questions dated after its cutoff, and labeling such questions for every year is costly. We address the first obstacle by comparing recipes that read only calls with recipes that start from an encoder of general text. We address the second obstacle with the calls themselves. Each analyst question is followed by an executive's answer, so the calls supply dated question-answer pairs without the need for labels, for training as well as for evaluation on calls held out from training.

\looseness=-1 From \NCallsIndex{} earnings calls held between September 2004 and September 2026, we train yearly vintages with cutoffs from 2010 to 2024, either from scratch on a fixed token budget or chained. A chained vintage continues the pretraining of the previous vintage on the new year's calls, which is continual pretraining on a chronologically ordered stream of text~\cite{jin2022lifelong}. Two further families start from ChronoBERT's vintages, one of them continued on our calls by domain-adaptive pretraining (DAPT)~\cite{gururangan2020dont}. We then apply contrastive fine-tuning to every vintage, on the question-answer pairs dated on or before its cutoff. We evaluate the vintages through retrieval on calls held out from pretraining and fine-tuning and through forecasts of returns and fundamentals. In both uses, we measure the look-ahead premium, which is the change in a score when a later vintage replaces the one available at the time.\footnote{\url{\CodeURL} holds the code, the call identifiers, and the result files and training records from which we build every table and every figure of results. The transcripts are licensed and are not redistributed.} We make three main contributions:
\begin{enumerate}
\item We build point-in-time retrievers for earnings calls and compare four training recipes. The vintages we train on calls alone, with \ParamsOurs M parameters, retrieve answers better than the general retrievers we tested, of up to 335M parameters, in every test quarter. An encoder with pretraining alone retrieves far worse than one with contrastive fine-tuning, and between our recipes, recall depends on how much of its window a vintage reads.
\item We show that point-in-time call embeddings forecast next-day returns beyond baselines built from past returns and firm size, in a design adapted from~\cite{he2026chronologically}. The embeddings raise the rank correlation of the baseline forecasts with next-day returns by about half, a statistically significant increase.
\item We measure look-ahead through the encoder and find that it depends on how the vintages are trained and on what the vectors are used for. A chained vintage that has read the year of the test quarter retrieves better than the vintage available at the time. A replay control that trains for the same number of tokens on text it has already read does not, so the gain comes from the later text. A later vintage of the family trained from scratch on a fixed budget shows no net gain, so the measured premium in retrieval depends on the recipe. A point-in-time language model that answers from the passages that a later chained vintage retrieves inherits the chained gain. The 2024 vintage of our families improves forecasts of fundamentals, although on calls that no vintage read the interval of its gain on the earnings surprise includes zero. Return forecasts with these families show no such improvement at any horizon we test, from the next day to the next quarter.
\end{enumerate}

In Sec.~\ref{sec:related}, we review related work. In Sec.~\ref{sec:data}, we describe the corpus and the point-in-time protocol, and in Sec.~\ref{sec:encoders} the four recipes. We report retrieval, forecasts and the look-ahead premium in Secs.~\ref{sec:retrieval} to~\ref{sec:lookahead} and conclude in Sec.~\ref{sec:conclusion}.

\section{Related Work}\label{sec:related}
\looseness=-1 DAPT continues the pretraining of a general encoder on text of the target domain before fine-tuning, and~\cite{gururangan2020dont} find that this second phase of pretraining leads to performance gains in four domains. In finance, FinEAS~\cite{gutierrez2022fineas} fine-tunes sentence embeddings for financial sentiment, and FinBERT~\cite{huang2023finbert} pretrains an encoder on filings, call transcripts and analyst reports. The finance retrievers discussed in Sec.~\ref{sec:intro} then train general encoders on queries or labels that are at least partly generated. On earnings calls,~\cite{juan2025cotrained} train a retriever of call content together with a generator of the questions analysts may ask. EvasionBench~\cite{ma2026evasionbench} extracts 22.7 million question-answer pairs from Capital IQ transcripts and trains a classifier of evasive answers on 84,000 labeled ones. We use such pairs as supervision for retrievers, dated so that each vintage trains only on pairs dated on or before its cutoff.

\looseness=-1 Look-ahead bias has been measured through the way a model is used. For instance,~\cite{gao2025detecting} estimate from a date-only recall query how likely a model is to know a firm's realized outcome, and find that forecasts from news headlines and from earnings calls gain predictive power where that likelihood is high. DatedGPT~\cite{yan2026datedgpt} finds a significant look-ahead premium when a prompted model forecasts returns from news headlines and was trained on text from the outcome period. Through frozen embeddings,~\cite{he2026chronologically} find no evidence that access to future information improves the performance of return portfolios in what they call their cleanest test for look-ahead bias. In that test, the vintage whose training text includes the test year replaces the one whose text ends the year before, so the two differ by one year of text. In retrieval-augmented systems,~\cite{zhao2026pointintime} make a passage eligible only if its timestamp precedes the prediction time, while the models that read the passages are not dated. In addition,~\cite{zhou2026historical} name a retriever's parameters as a channel of hindsight leakage and keep them fixed in their experiments.

\begin{figure*}[t]
\centering
\begin{tikzpicture}[
  font=\footnotesize,
  box/.style={draw=black!60, rounded corners=2pt, align=center, inner sep=2.5pt, minimum height=8mm},
  data/.style={box, fill=black!4},
  model/.style={box, fill=blue!6, text width=25mm},
  use/.style={box, fill=orange!8, text width=29mm},
  arr/.style={-{Stealth[length=1.8mm]}, black!70},
  node distance=3mm and 5.5mm]

\node[data, text width=19mm] (corpus) {Earnings calls\\ 2004 to 2026\\ (\NCallsIndex{} calls)};
\node[data, text width=21mm, right=of corpus] (windows) {Trainable calls\\ dated on or before\\ 31 Dec of $Y$};
\node[model, right=7mm of windows, yshift=13.5mm] (scratch) {From scratch,\\ 1.2B tokens each};
\node[model, below=1.5mm of scratch] (chain) {Chained: vintage $Y{-}1$\\ continued on year $Y$};
\node[model, below=1.5mm of chain] (abl) {ChronoBERT vintage $Y$\\ (general text)};
\node[model, below=1.5mm of abl] (dapt) {ChronoBERT $Y$ + DAPT\\ (2010, 2017, 2024)};
\node[box, fit=(scratch)(dapt), inner sep=2mm, draw=black!40, dashed,
      label={[font=\footnotesize]above:Pretraining recipes}] (recipes) {};
\node[box, text width=19mm, right=7mm of recipes.east |- windows] (task) {Contrastive\\ fine-tuning on\\ pairs dated\\ on or before $Y$};
\node[box, text width=19mm, right=of task] (reg) {Registry:\\ cutoff $Y$,\\ query date check};
\node[use, right=7mm of reg, yshift=10mm] (retr) {Retrieval on\\ held-out questions};
\node[use, below=1.5mm of retr] (fc) {Forecasts: ridge on\\ vectors, nearest neighbors};
\node[use, below=1.5mm of fc] (la) {Look-ahead: a later\\ vintage replaces $Y$};

\node[data, text width=21mm, below=15mm of windows] (pairs) {Analyst-question,\\ executive-answer pairs};
\node[data, text width=19mm, below=15mm of corpus] (held) {10\% of calls held\\ out by a hash\\ of the call id};

\draw[arr] (corpus) -- (windows);
\draw[arr] (corpus) -- (held);
\draw[arr] (windows.east) -- ++(2.5mm,0) |- (scratch.west);
\draw[arr] (windows.east) -- ++(2.5mm,0) |- (chain.west);
\draw[arr] (windows.east) -- ++(2.5mm,0) |- (dapt.west);
\draw[arr] (windows) -- (pairs);
\draw[arr] (recipes.east |- task) -- (task);
\draw[arr] (pairs.south) -- ++(0,-1.5mm) -| (task.south);
\draw[arr] (task) -- (reg);
\draw[arr] (reg.east) -- ++(3.5mm,0) |- (fc.west);
\draw[arr] (reg.east) -- ++(3.5mm,0) |- (la.west);
\draw[arr] (reg.east) -- ++(3.5mm,0) |- (retr.west);
\draw[arr] (held.south) -- ++(0,-3mm) -| ($(retr.east)+(3mm,0)$) -- (retr.east);
\end{tikzpicture}
\caption{From the corpus to the uses of the vintages. For each cutoff year $Y$ from 2010 to 2024, every pretraining recipe reads only text dated on or before 31 December of $Y$. Our two families read trainable calls, ChronoBERT reads general text, and ChronoBERT + DAPT reads both, at 2010, 2017 and 2024 only. Contrastive fine-tuning trains on question-answer pairs from the trainable calls, and the registry records the cutoff so that a query dated before it cannot use the vintage. Held-out calls supply every retrieval question and enter no stage of training other than the fit of the 2010 tokenizer.}
\label{fig:pipeline}
\end{figure*}
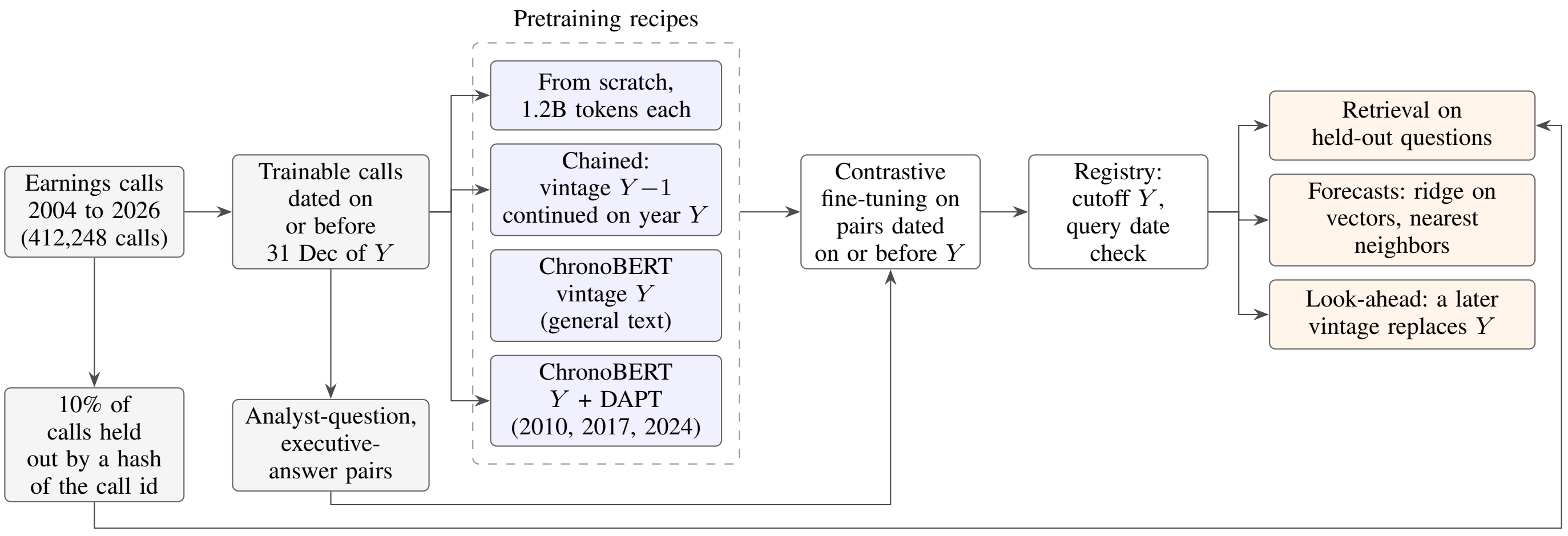
 
\looseness=-1 Point-in-time models also differ in how they train a sequence of vintages. DatedGPT trains each yearly model from scratch on an equal budget, which is our first recipe. In contrast,~\cite{kelly2026scaling} train decoders on one trillion chronologically filtered tokens in a single run and release monthly checkpoints. Without a yearly pretraining run,~\cite{lee2026palm} fit a low-rank adapter on text published before each decision date and leave the pretrained weights unchanged. ChronoBERT~\cite{he2026chronologically} is the closest work. Each of its yearly encoders continues the previous one on the new year's text at a constant learning rate, which is our second recipe. Its chain reads general text, while ours reads earnings calls, the text our tests are about. To that design, we add contrastive fine-tuning for retrieval, a family trained from scratch on domain text, and a replay control that distinguishes later text from longer training.

\section{Corpus and the Point-in-Time Protocol}\label{sec:data}
\looseness=-1 A vintage is point in time only if every text it reads carries a date, and only if the firms in the corpus do not depend on which firms survived. We take every earnings call in the Capital IQ transcripts on Wharton Research Data Services (WRDS), \NCallsIndex{} calls of more than 16,000 firms from September 2004 to September 2026, and keep every firm, whether or not it is still listed. Our vintages read calls through 2024, and our tests use calls through 2025. Capital IQ often holds several versions of one call, and no collection type is consistently the most complete, so we keep the version with the most text.\footnote{Capital IQ holds 3.2 transcripts per call on average, of collection types such as edited and audited copies, and revises them for weeks after the call. The version we keep can thus postdate the cutoff of a vintage that reads it, although it transcribes the same call, whose content was public once the call ended. We keep such a late version for \VersionLateMin\% to \VersionLateMax\% of the calls a vintage can train on at each cutoff from 2011 to 2024, and \VersionLateTen\% at the 2010 cutoff.} We date a call by its event date in Capital IQ. Membership in the S\&P~500 at the date of each call comes from the index history of the Center for Research in Security Prices (CRSP), and we use it only for the calls with fundamental targets in Sec.~\ref{sec:forecasting}.

\looseness=-1 A call is a sequence of speaker turns, a presentation by executives followed by questions from analysts and answers from executives. We split each call into chunks of about 2,400 characters, or about 500 tokens, and break a longer turn at a paragraph or sentence boundary. The chunks are the units that a vintage reads in pretraining and that we average into a call vector. In the question-and-answer section, a chunk starts at an analyst's question and holds the answers that follow it. We drop the turns of the call's operator, who introduces each analyst, unless a turn has at least 1,000 characters, since Capital IQ sometimes attributes a whole call to the operator. Each analyst question followed directly by an executive's turn gives one question-answer pair, and we keep a pair only if both turns have at least 120 characters, so we drop short turns such as thanks and greetings. We also take pairs only from calls that offer at least 5 answers of that length.

\looseness=-1 We hold out 10\% of calls, each in its entirety, from pretraining and contrastive fine-tuning.\footnote{We hold out a call when the first four bytes of the SHA-1 hash of its Capital IQ event identifier, prefixed with the string \texttt{ecalls-holdout-v1:}, fall in the lowest tenth of their range as a big-endian integer. The rule does not need a stored list, so anyone with the identifiers can reproduce the split.} The split is by call, so no chunk and no pair of a held-out call reaches any vintage, and every retrieval question comes from these calls.

\looseness=-1 The window of cutoff $Y$ holds the trainable calls, those not held out, dated on or before 31 December of $Y$, and vintage $Y$ reads only its window. Whether a call is held out depends only on its identifier, so the held-out calls are the same for every vintage and are about 10\% of the calls dated on or before each cutoff. We train a vintage for every cutoff from 2010 to 2024 (Fig.~\ref{fig:pipeline}), and we list the corpus at each cutoff in Table~\ref{tab:corpus}. We record each vintage with its cutoff in a registry and enforce two rules at query time: a query dated $d$ sees only calls dated on or before $d$, and a query that names a vintage with a later cutoff fails. For a test quarter in year $Y+1$, the vintage available at the time is therefore vintage $Y$.

\begin{table*}[t]
\centering
\footnotesize
\setlength{\tabcolsep}{3.2pt}
\caption{The corpus by cutoff year: the calls dated on or before 31 December of each cutoff year. Calls and chunks include the held-out calls, while the tokens for pretraining and the question-answer pairs for contrastive fine-tuning exclude them. Calls are in thousands (k), chunks and pairs in millions (M), and tokens in billions (B).}
\label{tab:corpus}
\begin{tabular}{l r r r r r r r r r r r r r r r}
\toprule
Cutoff & 2010 & 2011 & 2012 & 2013 & 2014 & 2015 & 2016 & 2017 & 2018 & 2019 & 2020 & 2021 & 2022 & 2023 & 2024 \\
\midrule
Calls (k) & $39.4$ & $57.4$ & $75.5$ & $93.4$ & $111.4$ & $129.4$ & $147.3$ & $167.6$ & $190.6$ & $214.6$ & $241.3$ & $269.2$ & $298.8$ & $328.8$ & $358.8$ \\
Held out (k) & $4.0$ & $5.8$ & $7.5$ & $9.3$ & $11.1$ & $13.0$ & $14.8$ & $16.8$ & $19.1$ & $21.5$ & $24.2$ & $27.0$ & $29.9$ & $32.8$ & $35.8$ \\
Chunks (M) & $1.53$ & $2.15$ & $2.76$ & $3.35$ & $3.91$ & $4.46$ & $4.98$ & $5.56$ & $6.20$ & $6.86$ & $7.62$ & $8.38$ & $9.17$ & $9.95$ & $10.72$ \\
Tokens (B) & $0.30$ & $0.44$ & $0.57$ & $0.71$ & $0.84$ & $0.98$ & $1.11$ & $1.25$ & $1.42$ & $1.58$ & $1.78$ & $1.99$ & $2.20$ & $2.41$ & $2.61$ \\
Pairs (M) & $0.49$ & $0.71$ & $0.93$ & $1.15$ & $1.37$ & $1.58$ & $1.79$ & $2.01$ & $2.25$ & $2.50$ & $2.76$ & $3.01$ & $3.28$ & $3.55$ & $3.81$ \\
\bottomrule
\end{tabular}
\end{table*}
\begin{table*}[t]
\centering
\footnotesize
\setlength{\tabcolsep}{2.7pt}
\caption{Retrievers and encoders compared. Parameters count the encoder body as loaded for the benchmark, and Dim.\ is the embedding dimension. Tokens are the tokens read in pretraining, and every from-scratch vintage reads the same 1.2B, independently of the size of its window. For our yearly families the range runs from the 2010 vintage to the 2024 one, and for ChronoBERT from its 1999 base, where its chain starts, to its 2024 vintage. FinBERT's entry is the size of its corpus, since the tokens it read are not stated. Contrastive fine-tuning trains on question-answer pairs from trainable calls dated on or before each vintage's cutoff. The replay control continues chained vintage $Y$ on its own last year for as many tokens as the next segment reads.}
\label{tab:families}
\begin{tabular}{l r r l l l l l}
\toprule
Model & Params (M) & Dim. & Pretraining text & Tokens & Tokenizer & Fine-tuning & Cutoff \\
\midrule
\multicolumn{8}{l}{\emph{References}} \\
BM25 & n/a & n/a & n/a & n/a & word counts & none & n/a \\
bge-small & 33.4 & 384 & general web & not stated & WordPiece & general pairs & unknown \\
bge-base & 109.5 & 768 & general web & not stated & WordPiece & general pairs & unknown \\
bge-large & 335.1 & 1024 & general web & not stated & WordPiece & general pairs & unknown \\
e5-base & 109.5 & 768 & general web & not stated & WordPiece & general pairs & unknown \\
FinBERT & 109.8 & 768 & filings, calls, reports & 4.9B corpus & FinVocab & none & corpus to 2019 \\
ChronoBERT & 149.0 & 768 & general & 460B to 525B & BPE, ModernBERT & none & 2010 to 2024 \\
\multicolumn{8}{l}{\emph{Contrastive fine-tuning on call question-answer pairs}} \\
ChronoBERT + contrastive fine-tuning & 149.0 & 768 & general & 460B to 525B & BPE, ModernBERT & call pairs & 2010 to 2024 \\
ChronoBERT + DAPT & 149.0 & 768 & general, then calls & +1.2B on calls & BPE, ModernBERT & call pairs & 2010, 2017, 2024 \\
Ours, from scratch & 33.8 & 384 & calls & 1.2B & BPE, 2010 calls & call pairs & 2010 to 2024 \\
Ours, chained & 33.8 & 384 & calls & 1.2B to 3.5B & BPE, 2010 calls & call pairs & 2010 to 2024 \\
\multicolumn{8}{l}{\emph{Controls for training length}} \\
Ours, twice the tokens & 33.8 & 384 & calls & 2.4B & BPE, 2010 calls & call pairs & 2024 \\
Ours, replay control & 33.8 & 384 & calls & +0.13B to +0.21B & BPE, 2010 calls & call pairs & 2010 to 2023 \\
\bottomrule
\end{tabular}
\end{table*}
 
\looseness=-1 For the forecasts of Sec.~\ref{sec:forecasting}, we use the calls of every firm that links to a stock in CRSP and add market data from WRDS. Daily returns, prices and share type and exchange flags come from the CRSP daily stock file, and the market return from the daily Fama-French factors. We link a call to its stock through the Capital IQ link to Compustat and the CRSP and Compustat link history, both valid on the call date. We compute the earnings surprise from quarterly earnings per share in Compustat's standard file, which includes restatements, for the fiscal quarter whose report date is within three days of the call. We convert Capital IQ's call start times from UTC to New York time.

\section{Point-in-Time Encoders}\label{sec:encoders}

We compare four recipes that differ in the encoder and in what it reads before contrastive fine-tuning, and we give every recipe the same fine-tuning, so a difference in retrieval comes from pretraining (Table~\ref{tab:families}).

\looseness=-1 Our encoder follows the ModernBERT architecture~\cite{warner2025smarter}, with 12 layers, a hidden size of 384 and \ParamsOurs M parameters. Its size is close to the 33.4M parameters of bge-small v1.5~\cite{xiao2024cpack}, so a difference from bge-small reflects how the model was trained rather than its size. It pretrains by masked language modeling and predicts 30\% of the tokens of each 768-token sequence, chosen at random. Every vintage uses one tokenizer, a byte-level byte-pair encoding (BPE) with 32,768 tokens fitted on every call dated on or before 2010, held-out calls included. These calls precede every cutoff and every test quarter, so the tokenizer reads no text after any cutoff. We split every digit into its own token, because with whole numbers, years such as 2008 and 2015 are single tokens while 2019 and 2024 are two, and a model could date a text by how its years split.

\looseness=-1 In the from-scratch family, we train each vintage $Y$ from random weights on window $Y$, for 1.2B tokens at every cutoff. With an equal budget, only the window varies from one vintage to the next, as in DatedGPT. At 1.2B tokens, the 2010 vintage reads its \TokensTwentyTen B tokens about four times, and~\cite{muennighoff2023scaling} find for decoder language models that up to 4 epochs of repeated data change the loss little against unique data. The 2024 vintage reads less than half of its window, so we also train a 2024 vintage on twice the tokens to measure what the budget costs.

\looseness=-1 The chained family follows the chain of ChronoBERT. Its 2010 vintage is the from-scratch one, and each later vintage $Y$ continues vintage $Y-1$ for one pass over the calls of year $Y$, so the chained 2024 vintage has read 3.5B tokens. ChronoBERT trains its chain at a constant learning rate, while we give each segment a fresh optimizer and its own warmup and decay. Every chained vintage thus ends its training at a tenth of its peak learning rate, as a from-scratch vintage does. To distinguish the effect of later text from that of longer training (Sec.~\ref{sec:lookahead}), a replay control continues each chained vintage $Y$ from 2010 to 2023 on the calls of year $Y$, which it has already read. The replay trains for as many tokens as the next segment reads, so we compare repeated with unique data at equal compute, as in~\cite{muennighoff2023scaling}.

\looseness=-1 The other two recipes start from the ChronoBERT vintage with the same cutoff, an encoder of \ParamsAbl M parameters pretrained on 460B to 525B tokens of general text. For ChronoBERT + contrastive fine-tuning, we apply our fine-tuning to the ChronoBERT vintage directly. To test whether domain text helps an encoder of fixed size, for ChronoBERT + DAPT we first continue its masked language modeling on our window for 1.2B tokens, with ChronoBERT's own tokenizer. We run this recipe at the 2010, 2017 and 2024 cutoffs.

\looseness=-1 In contrastive fine-tuning, we turn an encoder into a retriever by training it on the question-answer pairs of trainable calls dated on or before the vintage's cutoff. We use an InfoNCE loss~\cite{oord2018representation} that raises the cosine similarity of a question with its own answer against the other answers in the batch. Each batch holds up to 16 pairs, drawn at random, from each of 32 calls, about 370 to 405 pairs on average depending on the cutoff. Each question thus faces up to 15 answers from its own call and the rest from other calls. The batch size sets the number of negatives, so we keep it the same on every GPU by gradient caching~\cite{gao2021scaling}, which assembles the gradient of the full batch from pieces that fit in memory. Every vintage of every recipe trains for 4,000 steps (Table~\ref{tab:settings}) and represents a text by the mean of its token vectors. As in pretraining, the fixed budget means that the 2010 vintage reads each pair about three times and the 2024 vintage two fifths of its pairs. Every run trains on one GPU of a shared cluster in 16-bit floating point,\footnote{Most runs drew an NVIDIA RTX 2080 Ti, on which a from-scratch vintage pretrains in about 20 hours, a chained segment in about one hour, and contrastive fine-tuning in about six.} and the vintages of a family share every setting.

\begin{table}[t]
\centering
\footnotesize
\setlength{\tabcolsep}{3pt}
\caption{Training settings. Pretraining covers our from-scratch vintages, and each chained segment uses the same settings for one pass over its year. DAPT continues a ChronoBERT vintage on our window. Contrastive fine-tuning is the same for every recipe, with questions cut at 256 tokens and answers at 512, and $\tau$ is the InfoNCE temperature. Every stage uses AdamW with $\beta=(0.9, 0.98)$, $\epsilon=10^{-6}$ and weight decay 0.01, gradients clipped at norm 1, a linear warmup over the stated share of steps, and a cosine decay to a tenth of the peak learning rate.}
\label{tab:settings}
\begin{tabular}{l l l l}
\toprule
Setting & Pretraining & DAPT & Fine-tuning \\
\midrule
Objective & masking, 30\% & masking, 30\% & InfoNCE, $\tau=0.05$ \\
Length & 1.2B tokens & 1.2B tokens & 4,000 steps \\
Batch & 98,304 tokens & 98,304 tokens & 32 calls $\times$ up to 16 pairs \\
Sequence & 768 tokens & 768 tokens & 256 and 512 tokens \\
Peak rate & $6\times 10^{-4}$ & $5\times 10^{-5}$ & $2\times 10^{-5}$ \\
Warmup & 6\% & 6\% & 5\% \\
\bottomrule
\end{tabular}
\end{table}
 
\section{Retrieval}\label{sec:retrieval}
\looseness=-1 To measure how well each vintage retrieves without letting it score text it was trained on, we draw every question from the held-out calls of Sec.~\ref{sec:data}, on which no vintage trained. For each cutoff $Y$, we sample 400 held-out calls from the second quarter of year $Y+1$ and draw 3,000 of their question-answer pairs at random. We thus score each vintage on calls after its cutoff, as it would be deployed, rather than on one quarter shared by every vintage, which would favor the vintages whose windows end closest to it. We take the second quarter to leave a quarter between the cutoff and the test calls and to avoid the year-end calls of the first quarter. A dense retriever ranks the candidate answers to a question by the cosine similarity of their vectors to the question's vector, and BM25~\cite{robertson2009bm25} by word overlap. Within-call recall@1 is the share of questions whose own answer ranks first among the executive turns of at least 120 characters in the question-and-answer section of the same call, where every candidate comes from the same firm. When a second executive also answers, ranking that turn first counts as a miss for every retriever alike. Cross-call recall@1 is the same share among those turns in every sampled call. Representations of call transcripts largely encode the identity of the firm~\cite{yu2025same}, so a retriever can succeed across calls by recognizing the firm. Within a call the firm is fixed, so we lead with the within-call task. A call offers a median of \PoolMedianLo{} to \PoolMedianHi{} candidate answers per question, depending on the quarter, so a random ranking reaches a within-call recall@1 of about \ChanceWithin. In both tasks we rank a correct answer tied with others below all of them. Every retriever answers the same sample of questions, so we compare two retrievers question by question and draw 95\% intervals by resampling calls within each quarter. These intervals reflect the sampling of calls, since each vintage is one training run.

\looseness=-1 We score the general retrievers on our task, which they were not trained on, so by comparing with them we measure what dated, in-domain contrastive training adds. Their training cutoffs are unknown, so on early quarters their encoders may have read later text, which would favor them. For this reason, the findings about training recipes rest on the comparisons in which we hold contrastive fine-tuning fixed.

\begin{figure}[t]
\centering
\includegraphics[width=\columnwidth]{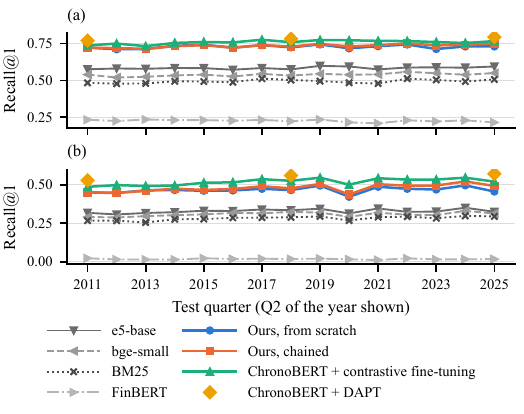}
\caption{Recall@1 on held-out calls by test quarter, for the vintage available at the time, so vintage $Y$ is scored on the second quarter of year $Y+1$. Each quarter has 3,000 analyst questions from 400 sampled calls. Panel (a): each question ranked against the executive answers in its own call. Panel (b): each question ranked against every answer of the sampled calls. Colored lines are retrievers with our contrastive fine-tuning, gray lines are references, and diamonds mark the three DAPT vintages (2010, 2017, 2024). FinBERT's corpus includes call transcripts through 2019. Every retriever is scored on the same random sample of calls and questions. bge-base, raw ChronoBERT and bge-large, the strongest general retriever across calls, are left out for legibility.}
\label{fig:recall}
\end{figure}

\looseness=-1 Our from-scratch vintages, with \ParamsOurs M parameters each (Table~\ref{tab:families}), reach a within-call recall@1 of \RecallOurs{} averaged over the 15 quarters, against \RecallEfive{} for e5-base~\cite{wang2022e5}, the strongest general retriever we tested, and \RecallBgeSmall{} for bge-small. The difference of \OursMinusEfive{} over e5-base is positive in every quarter (Fig.~\ref{fig:recall}), and across calls it is \OursMinusEfiveCross{} over e5-base and \OursMinusBestCross{} over bge-large, the strongest general retriever on that task. Recall@5, the share of questions whose own answer ranks among the first five, also credits a question whose answer ranks behind a second executive's turn. It is \RecallFiveOurs{} for our from-scratch vintages against \RecallFiveEfive{} for e5-base, and higher in every quarter. An encoder with pretraining alone retrieves poorly, because it learns to predict masked words and nothing in that objective places a question near its answer, which is what we train in contrastive fine-tuning. Raw ChronoBERT reaches \RecallChrono{} and FinBERT \RecallFinbert{}, above chance but far below every retriever with a contrastive stage, although FinBERT's pretraining corpus includes call transcripts through 2019. With contrastive fine-tuning alone, ChronoBERT gains \TaskStageGain{} (Table~\ref{tab:decomposition}).

\begin{table}[t]
\centering
\footnotesize
\setlength{\tabcolsep}{2pt}
\caption{Paired differences in recall@1 between retrievers, on the same held-out questions, averaged over the $Q$ test quarters both were scored on. Within is within-call and Across is cross-call recall@1. The 95\% interval of the within-call difference comes from resampling calls within each quarter. Each retriever is the vintage available at the time, and ChronoBERT has no contrastive fine-tuning. FinBERT is compared from 2020Q2, which its corpus cannot contain, DAPT on the 2010, 2017 and 2024 vintages, and the chain from its 2011 vintage, on 2012Q2. The two token budgets compare 2024 vintages on every quarter. The last two rows give the range across three seeds of contrastive fine-tuning, with pretraining fixed, of our from-scratch recall averaged over the 2010, 2017 and 2024 vintages, and the largest range of one vintage.}
\label{tab:decomposition}
\begin{tabular}{>{\raggedright\arraybackslash}p{31mm} r r c r}
\toprule
Difference & $Q$ & Within & 95\% interval & Across \\
\midrule
Ours, from scratch, minus e5-base & 15 & $+0.144$ & [$+0.139$, $+0.149$] & $+0.137$ \\
Ours, from scratch, minus FinBERT & 6 & $+0.508$ & [$+0.499$, $+0.518$] & $+0.451$ \\
ChronoBERT + contrastive fine-tuning minus ChronoBERT & 15 & $+0.559$ & [$+0.554$, $+0.564$] & $+0.501$ \\
ChronoBERT + DAPT minus ChronoBERT + contrastive fine-tuning & 3 & $+0.028$ & [$+0.021$, $+0.034$] & $+0.042$ \\
Ours, from scratch, minus ChronoBERT + contrastive fine-tuning & 15 & $-0.031$ & [$-0.035$, $-0.028$] & $-0.053$ \\
Ours, chained, minus ours, from scratch & 14 & $+0.007$ & [$+0.004$, $+0.010$] & $+0.014$ \\
Ours, twice the tokens, minus ours, from scratch & 15 & $+0.018$ & [$+0.015$, $+0.021$] & $+0.028$ \\
Training seed, range & n/a & $0.001$ & n/a & n/a \\
Training seed, range per vintage & n/a & $0.006$ & n/a & n/a \\
\bottomrule
\end{tabular}
\end{table}
 
\looseness=-1 Among the retrievers with our contrastive fine-tuning, the size of the pretrained encoder and the amount of text it read matter about as much as the domain of that text. ChronoBERT + contrastive fine-tuning scores \OursMinusAbl{} above our from-scratch vintages, with a larger encoder pretrained on about 400 times as many tokens of general text. Domain text still helps when we add it to the same encoder, since DAPT on our window raises the recall of ChronoBERT + contrastive fine-tuning by \DaptMinusAbl{} on the three vintages where we ran it. Between our two recipes, what matters is how much of its window a vintage reads. The chained vintages score above the from-scratch ones by a margin that grows in later quarters, where a chained vintage has read more tokens in total. A 2024 vintage on twice the tokens scores above the from-scratch one in every quarter, and on 2025Q2, after every cutoff, it reaches \StaleTwiceTwentyFour{}, close to the \StaleChainTwentyFour{} of the chained one. Averaged over the quarters from 2012Q2, the chained vintages score \ChainMinusScratch{} above the from-scratch ones, whose 2010 vintage they share. Averaged over all 15 quarters, the 2024 vintage on twice the tokens scores \TwiceGain{} above its from-scratch counterpart. Both averages exceed \SeedRangeVintage, the largest range of a from-scratch vintage's recall across three seeds of contrastive fine-tuning (Table~\ref{tab:decomposition}), although these seeds leave pretraining fixed.

\begin{table*}[t]
\centering
\footnotesize
\setlength{\tabcolsep}{4pt}
\caption{Call text in forecasts of returns and earnings surprises, with every vector from the vintage available at the time. \textbf{Panel A}: quarterly rank information coefficient (IC) of a ridge forecast refit each January on the price baselines and one text source, with the $t$ statistic over the test quarters from 2013 to 2025 in parentheses. The first row is the IC of the baselines alone, and each other row is the increment in IC when that text source is added. Next day and 5 days are the returns after entry at the first close at least an hour after the call starts, and next quarter is the return over trading days $+2$ to $+61$, each minus the market's. Session after is the raw open-to-close return of the first session after a call held before 08:30 or from 16:00 New York time. Sharpe ratios are annualized levels before trading costs, for the baselines alone in the first row and with text in the others. They come from an equal-weighted decile long-short of the calls entering each day, at the close or, for the session after, at the open, held one day on days with at least 20 such calls. A day without a position counts as zero return. Text alone is the next-day IC without the baselines. \textbf{Panel B}: IC of the mean outcome of the 50 most similar earlier calls of other firms, for the next-day return and, on S\&P 500 calls, the next standardized earnings surprise. The largest value in each column is in bold, among the text rows for an IC and among all rows for a Sharpe ratio.}
\label{tab:forecasts}
\begin{tabular}{l r r r r r r r}
\toprule
 & \multicolumn{4}{c}{IC, or increment in IC ($t$)} & \multicolumn{2}{c}{Sharpe ratio} & Text alone \\
\cmidrule(lr){2-5}\cmidrule(lr){6-7}\cmidrule(lr){8-8}
Panel A & Next day & 5 days & Session after & Next quarter & Next day & Session after & Next day \\
\midrule
Price baselines, IC & $0.025$ ($4.5$) & $0.026$ ($3.6$) & $0.027$ ($4.3$) & $0.011$ ($1.0$) & $1.73$ & $0.73$ &  \\
\addlinespace
\multicolumn{8}{l}{\emph{Text added to the baselines: increment in IC}} \\
Dictionary tone & $+0.007$ ($3.2$) & $+0.008$ ($2.6$) & $+0.022$ ($5.0$) & $+0.008$ ($1.7$) & $1.75$ & $1.65$ & $0.027$ ($7.7$) \\
TF-IDF & $+0.004$ ($1.0$) & $+0.012$ ($1.7$) & $+0.004$ ($0.8$) & $+0.010$ ($0.6$) & $1.57$ & $0.71$ & $0.020$ ($5.1$) \\
Ours, from scratch & $+0.012$ ($3.0$) & $+0.017$ ($2.8$) & $+0.023$ ($4.4$) & $+0.014$ ($0.9$) & $1.67$ & $1.59$ & $0.032$ ($8.1$) \\
Ours, chained & $+0.012$ ($3.2$) & {\boldmath$+0.021$} ($3.3$) & $+0.021$ ($3.7$) & $+0.018$ ($1.1$) & $1.57$ & $1.53$ & $0.033$ ($8.3$) \\
ChronoBERT + contrastive fine-tuning & {\boldmath$+0.013$} ($3.3$) & $+0.019$ ($2.7$) & {\boldmath$+0.029$} ($5.3$) & {\boldmath$+0.026$} ($1.7$) & {\boldmath$1.92$} & {\boldmath$1.82$} & {\boldmath$0.034$} ($9.2$) \\
ChronoBERT & $+0.008$ ($2.2$) & $+0.012$ ($2.0$) & $+0.018$ ($3.3$) & $+0.019$ ($1.3$) & $1.75$ & $1.58$ & $0.026$ ($6.9$) \\
\midrule
Panel B, IC & Next day & Surprise & & & & & \\
\midrule
Ours, from scratch & $0.011$ ($2.9$) & $0.048$ ($3.3$) & & & & & \\
Ours, chained & $0.011$ ($3.7$) & $0.049$ ($3.4$) & & & & & \\
ChronoBERT + contrastive fine-tuning & {\boldmath$0.012$} ($3.3$) & {\boldmath$0.056$} ($3.7$) & & & & & \\
ChronoBERT & $0.007$ ($2.3$) & $0.027$ ($2.1$) & & & & & \\
\bottomrule
\end{tabular}
\end{table*}
 
\section{Forecasting with Point-in-Time Embeddings}\label{sec:forecasting}
\looseness=-1 In our primary test, we compare the rank correlation with next-day returns of forecasts from text plus price baselines with that of baselines alone, in a design adapted from~\cite{he2026chronologically} to calls. The universe is every call since 2011 whose firm links, on the call date, to a common stock listed on NYSE, AMEX or Nasdaq. A call also needs a price of at least \$5 at the last close before it, so firms enter and leave the sample. Since a call lasts about an hour, we enter a position at the close of the call date for a call starting by 15:00 New York time, and at the next trading day's close otherwise. The primary target is the return over the trading day after entry minus the value-weighted market return. We also forecast, minus the market's, the 5-day return after entry and the return over trading days $+2$ to $+61$, where day 0 is the first trading day that closes after the call starts. For a call held before 08:30 or from 16:00, we add the raw open-to-close return of the first session after the call, which ends at the entry close. The price baselines are the market-adjusted reaction from the last close before the call to the entry close and the return over the week before the call. They also include momentum from twelve months to one month before the call and log market capitalization. For the next-quarter return, the announcement return over days $-1$ to $+1$ replaces the reaction, and the return over days $-21$ to $-2$ replaces that of the week. For the session after the call, the overnight gap to the session's open replaces the reaction, which would contain the session itself.

\looseness=-1 We use two fundamental targets to measure whether text describes the firm's next quarter, which a model trained on later text could know in advance. Both vary from one call to the next within a firm, and we compute them on the S\&P~500 calls. The first is the standardized earnings surprise of the next fiscal quarter, which is the change in earnings per share from the same quarter a year earlier, scaled by price~\cite{livnat2006comparing}. The second is abnormal volatility, after the text-based volatility forecasts of~\cite{kogan2009predicting}. Abnormal volatility is the log ratio of the realized variance of daily returns on days $+2$ to $+21$ to that on days $-21$ to $-2$. For the earnings surprise, the baselines are the current surprise, which is that of the quarter the call reports, the announcement return, momentum and market capitalization. For abnormal volatility, they are the log realized variances of the stock on days $-6$ to $-2$, $-21$ to $-2$ and $-252$ to $-22$, after the heterogeneous autoregressive model of~\cite{corsi2009simple}. They also include that of the market on days $-21$ to $-2$, the announcement return, its absolute value and market capitalization.

\looseness=-1 Each January from 2013, so that the first fit trains on two years of calls, we fit a ridge regression on every earlier call whose target was known before January 1. We choose the penalty among the six powers of ten from 0.1 to $10^4$ by cross-validation over four time-ordered folds of these training calls and winsorize the target at the 1st and 99th percentiles of the training rows. The vector of a call is the mean of its chunk vectors, scaled to unit length, and the encoders with our contrastive fine-tuning read the first 512 tokens of each chunk. Vintage $Y$, available at the time for test year $Y+1$, embeds the training and test calls alike. A call enters the regression through the first 32 principal components of its vector, fitted on the training calls, so every encoder adds the same number of predictors, independently of its dimension. Besides the encoders, the text sources are dictionary tone, the share of a call's words that are negative minus the share that are positive in the dictionary of~\cite{loughran2011liability}, and term frequency-inverse document frequency (TF-IDF). TF-IDF enters as the first 32 singular components of the matrix of the calls' words, fitted each January on calls whose return was known before the test year. We evaluate a forecast by the Spearman rank correlation between forecasts and realized targets across the calls of a calendar quarter, which finance calls the information coefficient (IC), a rank measure that suits forecasts used to sort calls. We average the IC over the 52 quarters from 2013 to 2025, 51 for the next-quarter return and the two fundamental targets, and compute its $t$ statistic across quarters.

\looseness=-1 At short horizons, the encoders with contrastive fine-tuning add a statistically significant increment to the IC of the price baselines (Table~\ref{tab:forecasts}). At the next day, and similarly over five days, our from-scratch vintages add \NextDayIncOurs{} ($t=\NextDayIncOursT$) to the baselines' IC of \NextDayBaseIC, about as much as the chained vintages and ChronoBERT + contrastive fine-tuning. Raw ChronoBERT adds less than our from-scratch vintages, by a margin that is significant at the next day but not over five days, and dictionary tone adds less than them by a margin that is not significant. In the first session after the call, dictionary tone adds about as much as the encoders, and at the next quarter text adds nothing significant. On the earnings surprise, the baselines reach an IC of \SueBaseIC{} and the text of our from-scratch vintages alone \SueAloneOursIC, but together they reach only \SueWithOursIC. On abnormal volatility, the baselines reach \AvBaseIC{} and the IC of the encoders alone is near zero. We therefore also fit each text source without the baselines.

\begin{figure*}[t]
\centering
\includegraphics[width=\textwidth]{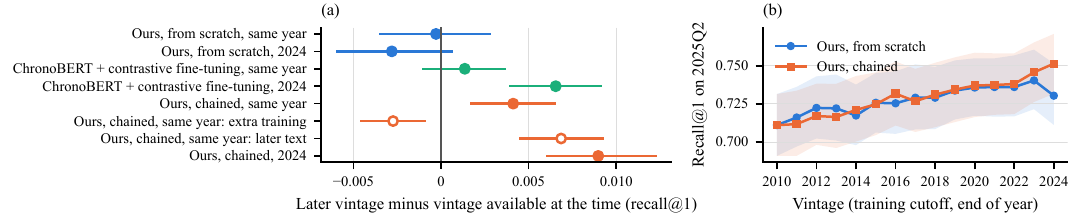}
\caption{Look-ahead through the encoder. \textbf{Panel (a)}: the premium of a later vintage over the vintage available at the time, in within-call recall@1 on the same held-out questions, averaged over the 14 test quarters 2011Q2 to 2024Q2, with 95\% intervals from resampling calls. The same year is vintage $Y+1$ on the test quarter of vintage $Y$, and 2024 is the 2024 vintage on every quarter. Hollow markers split the chain's same-year premium into extra training (replay minus vintage $Y$) and later text (vintage $Y+1$ minus replay), where the replay control continues chained vintage $Y$ on year $Y$ for as many tokens as the next segment reads. \textbf{Panel (b)}: the recall@1 on 2025Q2, a quarter after every cutoff, of each vintage of our two families, which share their 2010 vintage, with shaded 95\% intervals.}
\label{fig:lookahead}
\end{figure*}
 \looseness=-1 As in~\cite{he2026chronologically}, we also trade the forecasts. On each trading day with at least 20 calls entering at its close, \NextDayShareTraded\% of days, we buy the tenth of those calls with the highest forecasts and sell the tenth with the lowest, equally weighted. We hold the positions for one day and report Sharpe ratios before trading costs. For the session after the call, we enter at the session's open, by which a call held before 08:30 has ended, and trade on the \OtcShareTraded\% of days with at least 20 such calls. At the next day the Sharpe ratio of the baselines, \NextDaySharpeBase, changes little when we add encoder vectors, while in the session after the call it more than doubles. Both legs turn over on every traded day, so the baselines break even at a one-way cost of \BreakEvenBase{} basis points at the next day, \BreakEvenOurs{} with our from-scratch vectors, and \BreakEvenOtcOurs{} with them in the session after the call.

\looseness=-1 From retrieval itself we build a nearest-neighbor forecast, which fits no parameters, so its IC depends only on which calls the vectors place close together. For each test call, we average the outcomes of the 50 most similar earlier calls of other firms, and with our from-scratch vintages 10 or 200 neighbors give about the same IC for the surprise. We use only calls whose outcome was known before the test call's event day, and vectors from the vintage available at the time, and we winsorize the outcomes as we do the regression target. With our from-scratch vintages its IC for the next surprise is \KnnSueOurs{} ($t=\KnnSueOursT$), close to ChronoBERT + contrastive fine-tuning and above raw ChronoBERT (Table~\ref{tab:forecasts}, Panel~B). Calls our retrievers place close together tend to precede similar surprises.

\section{Look-Ahead through the Encoder}\label{sec:lookahead}
\looseness=-1 A later vintage has read text written after the date of a test, and that text can describe the outcomes the test asks about. We compute the look-ahead premium of Sec.~\ref{sec:intro} on the same questions or the same forecasting rows, for two later vintages (Table~\ref{tab:premia}). The 2024 vintage replaces the vintage available at the time in every test quarter. The same-year vintage, vintage $Y+1$ on the test calls of vintage $Y$, replaces it as in the test of~\cite{he2026chronologically} described in Sec.~\ref{sec:related}. The premium is a candidate measure of look-ahead, because a later vintage can also differ in how long it trained, which we control for below.

\looseness=-1 In retrieval, a later from-scratch vintage shows no net gain. Its premium is \ScratchSameYear{} in the same-year test and \ScratchLate{} with the 2024 vintage, and both intervals include zero (Fig.~\ref{fig:lookahead}a), as do those of two further 2024 vintages trained with other seeds of contrastive fine-tuning. The chained family, which we train at one seed, does gain, \ChainSameYear{} [\ChainSameYearLo, \ChainSameYearHi] in the same-year test, in \ChainSameYearPos{} of 14 quarters, and \ChainLate{} with the 2024 vintage. ChronoBERT + contrastive fine-tuning, whose pretrained encoders are also trained as a chain, gains \AblLate{} with the 2024 vintage, as the chained family does, with an interval above zero at all three seeds.

\begin{table*}[t]
\centering
\footnotesize
\setlength{\tabcolsep}{4pt}
\caption{Look-ahead premium across uses of the vectors: a later vintage's score minus that of the vintage available at the time. Retrieval is within-call recall@1 on held-out questions averaged over the 14 quarters 2011Q2 to 2024Q2, and every other column averages the 48 quarters 2013Q1 to 2024Q4. Nearest neighbors is the forecast of Table~\ref{tab:forecasts}, Panel B. Forecasts are the change in quarterly rank IC of text and baselines, for three return targets of Table~\ref{tab:forecasts} and, on S\&P 500 calls, the next standardized earnings surprise and the abnormal volatility after the call. \textbf{Panel A}: the later vintage is the 2024 one. \textbf{Panel B}: that of the test year. Raw ChronoBERT has no contrastive stage and no retrieval premium. Bold values have a 95\% interval that excludes zero, from resampling calls for retrieval and a moving-block bootstrap over quarters otherwise.}
\label{tab:premia}
\begin{tabular}{l r r r r r r r r}
\toprule
 & & \multicolumn{2}{c}{Nearest neighbors} & \multicolumn{5}{c}{Forecast with baselines} \\
\cmidrule(lr){3-4}\cmidrule(lr){5-9}
 & Retrieval & Next day & Surprise & Next day & Session after & Next quarter & Surprise & Volatility \\
\midrule
\multicolumn{9}{l}{\emph{Panel A: the 2024 vintage}} \\
Ours, from scratch & $-0.0028$ & $+0.0006$ & $+0.0032$ & {\boldmath$-0.0023$} & $-0.0012$ & $+0.0001$ & {\boldmath$+0.0044$} & {\boldmath$+0.0044$} \\
Ours, chained & {\boldmath$+0.0090$} & $-0.0008$ & $+0.0085$ & {\boldmath$-0.0022$} & $0.0000$ & $-0.0017$ & {\boldmath$+0.0034$} & {\boldmath$+0.0068$} \\
ChronoBERT + contrastive fine-tuning & {\boldmath$+0.0065$} & $+0.0025$ & $+0.0005$ & $-0.0016$ & $+0.0008$ & {\boldmath$-0.0042$} & $-0.0010$ & $-0.0030$ \\
ChronoBERT & n/a & $-0.0012$ & $+0.0008$ & $+0.0017$ & $+0.0016$ & $+0.0044$ & {\boldmath$+0.0059$} & $-0.0005$ \\
\addlinespace
\multicolumn{9}{l}{\emph{Panel B: the same-year vintage}} \\
Ours, from scratch & $-0.0003$ & $-0.0006$ & $-0.0024$ & $+0.0001$ & $-0.0014$ & $+0.0016$ & $-0.0005$ & $+0.0008$ \\
Ours, chained & {\boldmath$+0.0041$} & $-0.0008$ & $+0.0016$ & $+0.0001$ & $-0.0005$ & $-0.0016$ & $+0.0010$ & {\boldmath$+0.0034$} \\
ChronoBERT + contrastive fine-tuning & $+0.0014$ & $+0.0014$ & {\boldmath$-0.0053$} & $+0.0008$ & $-0.0006$ & $-0.0013$ & $-0.0014$ & $-0.0011$ \\
ChronoBERT & n/a & $+0.0026$ & $-0.0010$ & {\boldmath$+0.0023$} & $+0.0002$ & $+0.0020$ & {\boldmath$+0.0020$} & $+0.0013$ \\
\bottomrule
\end{tabular}
\end{table*}
 
\looseness=-1 The chained gain can come from the text of year $Y+1$, which is look-ahead, or from the extra training of each chained segment, which we isolate with the replay control of Sec.~\ref{sec:encoders}. Extra training alone does not raise recall, since the replay scores \ReplayCompute{} against vintage $Y$ and is higher in only \ReplayComputeWins{} of 14 quarters. The later text does, since vintage $Y+1$ scores \ReplayLater{} [\ReplayLaterLo, \ReplayLaterHi] above the replay, in \ReplayLaterWins{} of 14 quarters. The replay is fine-tuned on the pairs of vintage $Y$, so the later text enters both pretraining and contrastive fine-tuning. We take the same-year premium as the conservative estimate of the leak, because the difference against the replay also contains the replay's own small loss, which may come from rereading text or from restarting the learning-rate schedule. Whether the later text helps because it describes the test quarter or because it is new, a backtest with the later vintage uses text unavailable at the time. The later text that matters is not mainly the test firm restating its answers. We split each quarter's questions into thirds by the largest TF-IDF cosine between the answer and the firm's next four calls in the later vintage's window, a measure of restatement in the same words. In the 2024 test, whose window holds more of those calls, the premium stays above zero on the least restated third, at \RestateLeakyLow{} [\RestateLeakyLowLo, \RestateLeakyLowHi], and the most restated third exceeds it by \RestateLeakyGap{} [\RestateLeakyGapLo, \RestateLeakyGapHi], an interval that includes zero.

\begin{table}[!tb]
\centering
\footnotesize
\setlength{\tabcolsep}{1.5pt}
\caption{Look-ahead through the retriever, for a language model that answers from what the retriever returns. For 1,000 held-out analyst questions in each of the 14 quarters 2011Q2 to 2024Q2, a retriever returns the 3 closest answers from the quarter's sampled held-out calls. ChronoGPT-Instruct with the cutoff year of the vintage available at the time reads them with the question. The score is its mean log-probability per token, in nats, of the first 300 tokens of the executive's real answer. Own answer is the share of questions whose real answer is among the 3 excerpts, under the vintage available at the time, vintage $Y$, and the 2024 vintage. The premium is the 2024 retriever's score minus that of the vintage available at the time, with 95\% intervals from resampling calls. The last row is the same test with each question's real answer removed from the candidates. Pos.\ counts the quarters with a positive premium, and bold premia have an interval that excludes zero.}
\label{tab:rag}
\begin{tabular}{>{\raggedright\arraybackslash}p{19.5mm} r r r c r}
\toprule
 & \multicolumn{2}{c}{Own answer (\%)} & & & \\
\cmidrule(lr){2-3}
Family & Vintage $Y$ & 2024 & Premium & 95\% interval & Pos. \\
\midrule
Ours, chained & $63.5$ & $65.8$ & {\boldmath$+0.0764$} & [$+0.0568$, $+0.0942$] & 13/14 \\
Ours, from scratch & $62.9$ & $63.1$ & $+0.0094$ & [$-0.0112$, $+0.0300$] & 8/14 \\
Ours, chained, own answer removed & n/a & n/a & $+0.0018$ & [$-0.0002$, $+0.0036$] & 12/14 \\
\bottomrule
\end{tabular}
\end{table}
 
\looseness=-1 The gain from later text also appears when we score every vintage on one quarter after all cutoffs, the comparison that~\cite{lee2026palm} make for decoder signals on returns. On 2025Q2, older chained vintages score lower by \StaleSlopeChain{} recall@1 per year of staleness, and the 2010 vintage scores \StaleChainTwentyTen{} below the 2024 one (Fig.~\ref{fig:lookahead}b), so a vintage that has read more recent calls retrieves better on a later quarter. The from-scratch family declines by \StaleSlopeScratch{} per year, but its 2024 vintage scores below its 2023 vintage at each of three seeds of contrastive fine-tuning, because a vintage on a fixed budget reads each year of a longer window less. The 95\% interval of that difference includes zero. In a premium test, the vintage available at the time has the shorter window and so rereads its years more often, which can offset a small leak in the later vintage. We therefore read the from-scratch result as no net gain rather than no leak.

\looseness=-1 The retrieval premium also reaches a language model that answers from what the retriever returns. For 1,000 of the held-out questions of Sec.~\ref{sec:retrieval} in each quarter, a retriever returns the 3 answers of the quarter's sampled held-out calls closest to the question. The candidates include answers from calls held later in the quarter, for both retrievers alike. ChronoGPT-Instruct~\cite{he2025instruction} with the cutoff year of the vintage available at the time then reads them with the question. The reader's training data end at its cutoff, and the vintage available at the time and the 2024 vintage feed the same reader, so only the retriever differs. We score the reader's mean log-probability per token, in nats, of the first 300 tokens of the executive's real answer. The excerpts of the chained vintage available at the time raise it from $-\RagClosedLevel$ without excerpts to $-\RagPitLevel$. Replacing the chained vintage available at the time with the 2024 one raises the score by a further \RagChain{} [\RagChainLo, \RagChainHi], in \RagChainPos{} of 14 quarters, while the from-scratch family shows no significant premium (Table~\ref{tab:rag}). The premium comes from retrieving the answer itself, which the 2024 chained vintage places among the excerpts for \RagGoldChainLate\% of questions against \RagGoldChainPit\% for the vintage available at the time. When we remove that answer from the candidates, the premium falls to \RagNoGold, with an interval that includes zero. The premium is therefore the part of the retriever's own leak that reaches the reader, and a backtest of a retrieval-augmented system inherits its retriever's leak when its corpus holds the passage that answers the question.

\looseness=-1 In forecasts, the premium depends on the target (Table~\ref{tab:premia}). On returns it is near zero or negative for every encoder with our contrastive fine-tuning, including the nearest-neighbor forecast, and raw ChronoBERT shows a positive premium whose interval excludes zero only at the next day in the same-year test. A model prompted as in DatedGPT and~\cite{gao2025detecting} can state what it read about an outcome, while a regression on frozen embeddings can use only what the vectors encode. On returns,~\cite{he2026chronologically} and~\cite{kelly2026scaling} also find little difference. On the two fundamental targets, the 2024 vintage raises the IC of both of our families, by \SueLateOurs{} ($t=\SueLateOursT$) on the earnings surprise for our from-scratch vintages, and it raises that of raw ChronoBERT on the surprise as well. On the earnings surprise, the premium with the text alone is larger, \SueLateOursAlone{} ($t=\SueLateOursAloneT$) for our from-scratch vintages and \SueLateChainAlone{} ($t=\SueLateChainAloneT$) for the chained ones. In the same-year test, of our two families only the chained one gains significantly, on abnormal volatility. Since most of the forecast calls were in the training text of the 2024 vintage, we also compute the premium on the held-out calls, which no vintage read. There the point estimate of the premium of the text alone is about as large, \SueAloneOursHeld{} against \SueAloneOursTrained{} on the trained calls for our from-scratch vintages, but on about \SueHeldPerQuarter{} calls per quarter its interval, [\SueAloneOursHeldLo, \SueAloneOursHeldHi], includes zero. The held-out calls are therefore consistent with a gain that does not require having read the test call, which would suggest that a later vintage carries knowledge of later quarters in general.

\looseness=-1 Since we compute \NForecastPremia{} forecast premia, for the four encoder families of Table~\ref{tab:premia}, six targets and two later vintages, each with and without the baselines, a single $t$ statistic overstates the evidence. With $t$ statistics that treat quarters as independent, \NBonferroni{} pass a Bonferroni bound of $|t| > \BonferroniBound$ under the normal approximation, all on the fundamental targets. With the moving-block bootstrap of Table~\ref{tab:premia}, \NBonferroniBlock{} pass, the surprise premia of the 2024 vintage with the text alone, at $t$ of \SueLateOursAloneBlockT{} from scratch and \SueLateChainAloneBlockT{} chained. Against the bound of \BonferroniBoundStudent{} from the $t$ distribution with 47 degrees of freedom, \NBonferroniStudent{} pass with independent quarters, both on the surprise, and only the chained one with the bootstrap. The evidence for look-ahead in forecasts therefore rests on the earnings surprise and on the pattern across targets and families. With penalties up to $10^7$, the from-scratch increments and 2024 premia for the next day, the session after the call and the surprise change by under 0.001.

\section{Conclusions}\label{sec:conclusion}
\looseness=-1 With contrastive fine-tuning on the calls' own question-answer pairs, we obtain retrievers that rank answers better than the general retrievers we tested in every test quarter, at the size of bge-small and with a known cutoff. Their call embeddings also raise the next-day IC of the price baselines by about half. Used live, such a retriever gains from retraining as calls arrive, since newer chained vintages retrieve better on a fixed quarter. For encoders of this size on call text, the vintage available at the time is the one to use in a backtest of retrieval, of a language model that reads retrieved passages, or of a forecast.

\looseness=-1 Our results come from one corpus with point-in-time encoders of at most \ParamsAbl M parameters, and our training and test pairs need the question-answer turns of a call, which most news articles, filings and broker reports lack. The ChronoBERT-based recipes keep ChronoBERT's tokenizer of 2024, which splits 2010 and 2024 call text into tokens at about the same rate. In the replay control we restart the learning-rate schedule as in our chain, so carrying its result to a chain trained at a constant rate, as ChronoBERT is, is an extrapolation. The gains from text in forecasts are small. Finally, we leave to future work whether the premium through the encoder grows with the size of the model, and whether the premium that reaches a language model grows with the size of the reader.

\bibliographystyle{IEEEtran}

\end{document}